# Which structure of academic articles do referees pay more attention to?: perspective of peer review and full-text of academic articles

Chenglei Qin[1] and Chengzhi Zhang[1,2,*]

1. School of Economics and Management, Nanjing University of Science and Technology, Nanjing, China
2. Key Laboratory of Rich-media Knowledge Organization and Service of Digital Publishing Content, Institute of Scientific and Technical Information of China, Beijing, China

**Abstract**
**Purpose:** The purpose of this paper is to explore which structures of academic articles referees would pay more attention to, what specific content referees focus on, and whether the distribution of PRC is related to the citations.
**Design/methodology/approach:** Firstly, utilizing the feature words of section title and hierarchical attention network model (HAN) to identify the academic article structures. Secondly, analyzing the distribution of PRC in different structures according to the position information extracted by rules in PRC. Thirdly, analyzing the distribution of feature words of PRC extracted by the Chi-square test and TF-IDF in different structures. Finally, four correlation analysis methods are used to analyze whether the distribution of PRC in different structures is correlated to the citations.
**Findings:** The count of PRC distributed in Materials & Methods and Results section is significantly more than that in the structure of Introduction and Discussion, indicating that referees pay more attention to the Material and Methods and Results. The distribution of feature words of PRC in different structures is obviously different, which can reflect the content of referees' concern. There is no correlation between the distribution of PRC in different structures and the citations.
**Research limitations:** Due to the differences in the way referees write peer review reports, the rules used to extract position information cannot cover all PRC.
**Originality/value:** The paper finds a pattern in the distribution of PRC in different academic article structures proving the long-term empirical understanding. It also provides insight into academic article writing: researchers should ensure the scientificity of methods and the reliability of results when writing academic article to obtain a high degree of recognition from referees.
**Paper type:** Research paper

## Introduction

The peer review mechanism plays a critical role in scientific communication and is the closest to the actual state of the evaluated object (Narin, 1978), which is a practical guarantee of scientific quality (Mark Ware Consulting, 2016; P€oschl, 2004). As a carrier for recording the peer review process, the review report records the evaluation of the manuscript, which contains rich information such as views, sentiment and expert knowledge. For a long time, due to the limitations of the traditional peer review mechanism, there has been no large-scale open access to peer review comments (PRC). And limited by the development of text mining technology, researchers cannot unveil the mystery of peer review from the perspective of text content.

Due to the lengthy review cycle and prejudice of the traditional peer review mechanism, the open peer review came into being further to improve the fairness and transparency of the review process. Some influential journals, well-known publishing groups and scientific research institutions began to adopt the open peer review as early as 2001. ACP journal (Atmospheric

Chemistry and Physics), belongs to the European Geosciences Union (EGU), opened the interaction among referees, authors and public. Then, the journal published the interaction content together with papers (P€oschl, 2004, 2012). As of December 2020, 19 journals of the EGU have adopted the peer review mechanism. In January 2016, Nature Communication began implementing a transparent review mechanism, publishing referee comments, author rebuttal letters and editor decisions [1]. The journal was an early signatory to the open letter Open letter on the publication of peer review reports [2] published by ASApbio (Accelerating Science and Publication in Biology) in 2018. To date, 377 journals have signed and executed the open letter. In May 2019, all PloS journals offered authors the option to publish their peer review history alongside their accepted manuscripts (Madison, 2019). In December 2019, Nature Research added eight new journals, including Nature and Nature Biomedical Engineering, to adopt the transparent peer review mechanism [3]. In addition, the OpenReview.net platform, founded in 2013, provides a configurable solution to peer review for conferences and journals to promote openness in scientific communication. As seen above, open peer review is a trend, and the open access of PRC provides a data basis for analyzing and mining the peer review.

With the rapid development of deep learning, text mining technology has significantly improved its performance in text representation, text classification and other text processing tasks. It is now time to conduct research on peer review from the perspective of text content. Three research questions will be explored in this paper:

RQ1. Which article structures do referee pay more attention to?
RQ2. What specific content do referees focus on in different structures?
RQ3. Is the distribution of PRC in different structures related to the citations?

We take the original manuscripts and PRC published in ACP from 2001 to 2016 as the research objects, and mainly utilize the hierarchical attention network (HAN) model (Yang et al., 2017) to identify the article structures. Then, we study the distribution of PRC in different structures of academic articles. Moreover, we combined the Chi-square test (Said et al., 2020) with TF-IDF (Singhal, 2001) to study the distribution of feature words of PRC in different structures. In addition, we employ the Spearman correlation coefficient, cumulative distribution function, K-S test and negative binomial regression to analyze whether the distribution of PRC in different structures correlates with the citations.

## Related work

The research in this paper mainly involves the recognition of academic article structures, the mining of PRC and the relationship between the distribution of PRC and the citations. This section briefly introduces the related work of these three aspects.

**Academic article structures recognition**
Academic articles of social sciences, natural sciences, engineering and computer science usually can be divided into four parts: Introduction, Materials & Methods, Results and Discussion, denoted as IMRaD or IMRD (Williams, 2018), and the taking shape can be found in the book *Etudes sur la Biere* published in 1876 (Day, 1989; Pasteur, 1876). The IMRaD structure intuitively reflects the process of scientific discovery (George Mason University Writing Center, 2014). The standardization structure of articles ensures the effective communication and dissemination of scientific discoveries in the academic community, which is convenient for readers to read papers from different perspectives and find relevant information from a specific location (Parlindungan, 2012). Wu (2011) has a similar view that scientific progress depends on a rigorous publishing process. The IMRaD structure can help authors organize content and help editors and referees evaluate manuscripts. For readers, the IMRaD structure can help locate specific information efficiently without browsing the entire paper. In addition, Teodosiu (2019) thinks that the IMRaD structure is the best form of expression for articles, and a clear and standardized structure helps to publish papers. As early as the 1970s, the American National Standard Institute (ANSI) and the International Committee of Medical Journal Editors (ICMJE) used the IMRaD structure as a standard, and it became the written format for most journals (International Committee of Medical Journal Editors, 1991)。

The recognition of the structural function of sentences in academic articles was concerned firstly by researchers. As early as 2003, McKnight and Srinivasan (2003) divided the sentences of 7,253 abstracts into four functions: Instruction, Method, Result and Conclusion. They used the bag-of-words model and trained classifiers to predict the function of sentences in unstructured abstracts. The experiments showed that the method achieved good results. Agarwal and Yu (2009) explored whether sentences can be divided into IMRaD structures in the biomedical field with a consistency of 82.14% annotation. They then tested the effect of rules-based, SVM and other methods to identify the sentence functions, and the paper found the performance of Naïve Bayes models was better than others. Nam *et al.* (2016) tested the effect of the bag-of-words, language features, grammatical features and structural features in the function recognition of summary sentences. They found that language features can contribute to classify sentence functions.

The current typical approach to identifying the academic article structures is to identify which structure the article section belongs to. Researchers can divide the article structures in some fields according to the section title. Hu *et al.* (2013) explored the distribution of citations in the body of scientific articles based on 350 full text articles from the *Journal of Informetrics*. They divided the articles into four parts based on the section title, namely, Introduction, Method, Results and Conclusions (IMRC). Zhang *et al*. (2021) explored the distribution of adverbs and adjectives used by reviewers in different structures of academic papers based on 3,329 review reports from the British Medical Journal. They divided the structures into seven parts by manual: Overall, Abstract, Introduction, Methods, Results, Discussion and Other.

Usually, if the structure of the paper is clear, it is not necessary to automatically identify the structure of academic articles, such as those from PLoS. However, there is a situation that the paper structure cannot be directly judged according to the section title. Ribeiro *et al.* (2018) thought that reading efficiency would be improved if researchers could identify the article structures and extract it to the readers who need part of the structures. They tested the performance of five classifiers in recognizing the 129 articles structures from the PubMed Central database. They found that the Voting Feature Intervals algorithm performed best, and the best result of the accuracy value is 71.38%. Ahmed and Afzal (2020) thought that term-based literature retrieval could not meet special needs. For example, the current retrieval system cannot return the literature containing a specific term (e.g., "PageRank") in the structure of the results. Traditional paper structure recognition methods based on keywords, paper templates and references ignore some valuable features. They utilized the features of in-text citation count, figure count and table count, and section subheadings to map the structural function of section to IMRaD based on 5,000 articles from CiteSeer. The $F_1$ value is 97.50% which is higher than several traditional methods. Ma et al. (2020) hold a similar view and thought that identifying the structure of academic papers was the basis for achieving accurate retrieval of academic papers. They found that the section title was more beneficial than the in-section text while identifying the article structure based on 3,992 articles from ACL. Furthermore, they found that the traditional logistic regression (LR) and support vector machine (SVM) performed better than deep learning models, with a Macro-$F_1$ value of 92.49%. Then Ma *et al.* (2022) combined contextual information of current section with deep learning models to identify the structure of academic articles, with a Macro-$F_1$ of 94.71%. There is an obvious defect in the academic article structure recognition methods mentioned above, that is, the experimental data source is single that only from one journal or one conference. The result of recognition depends on the standardization of the articles structure—the more standardized, the higher the accuracy. Moreover, it can be found that the recognition accuracy of Introduction and Materials & Methods is higher than other structure from the existing work. The high F value contributes from the Introduction and Materials & Methods is small. Qin and Zhang (2020) did a comprehensive study. They compared the effects of traditional machine learning models, basic deep learning models, Bert (Devlin *et al.*, 2018) models and HAN models with different granularities on the recognition of academic articles structure in different datasets. The experimental results showed that the sentence-level HAN model was the best. The all Macro-$F_1$ values were higher than 0.98 in the dataset with standardized structures (PLoS Biology, Medicine, Genetics and Computational Biology), and the highest F1 value in Atmospheric Chemistry and Physics was 0.8661.

**Mining of peer review comments**
With the continuous improvement of open access to peer review reports, research related to the mining of PRC has attracted increasing attention from the academic community. Researchers mainly focused on evaluating the quality of PRC, predicting the accepted score of papers based on PRC, extracting the arguments and argument pairs from PRC and conducting relevant research on the combination of PRC and scientific metrics.

In the quality assessment of PRC, Ausloos *et al.* (2016) gathered the statistics on the count,

variation and distribution of words in PRC in the *Journal of the Serbian Chemical Society*, and the quality of PRC was quantified by using Zipf law. Ramachandran *et al.* (2017) used NLP technology to measure the quality of PRC based on six aspects: the type of PRC, the relevance of PRC, the coverage rate of PRC, the tone of PRC, the count of PRC and whether to plagiarize, which can provide feedback to the referees promptly.

Predicting the recommendation score of the manuscript according to PRC can assist the editor in making decisions. Wang and Wan (2018) proposed a multi-sample learning network model based on the abstract memory mechanism, used peer-reviewed data from the ICLR conference to calculate the sentiment of PRC and predicted the final state of the manuscript. Ghosal *et al.* (2019a, 2019b) used the convolutional neural network model to represent sentences in PRC and utilized the open-source tool VADER sentiment analyzer (Hutto and Gilbert, 2014) to calculate the sentiment polarity of PRC. They finally employed a multilayer perceptron to predict the recommendation score of the manuscript. Li *et al.* (2020) proposed a multi-task shared structure coding method to predict the scores of six aspects of PRC from the ICLR and ACL conference, such as clarity, originality and accuracy, which had better performance than the single-task method and the Naïve multi-task method.

The extraction of arguments and argument pairs in PRC has attracted the attention of researchers. Hua *et al.* (2019) annotated 400 peer review reports from the four conferences of ICLR, ACL, UAI and NeurIPS with evaluation, request, fact, reference, quote and non-arg. They built a classification model, predicted the types of PRC and analyzed different arguments distribution in the four conferences. However, the study also had limitations that the conference field and the review requirements may affect the conclusions. Cheng *et al.* (2020) thought there was a rich interactive and controversial discussion between PRC and author rebuttals. They proposed a multi-task learning framework based on a hierarchical LSTM network to extract argument pairs.

PRC combined with scientific metrics can predict the citations of articles. Li *et al.* (2019) thought that predicting the citations of papers is an important research task to assess the future influence of an academic paper automatically. The jointed width and depth component model was proposed, and the PRC was represented with a matching mechanism to predict the citations. Zong *et al.* (2020) explored whether the sentiment polarity of post-publication PRC from PubPeer, F1000 and ResearchGate affected the citations through content analysis method, control variable method, negative binomial regression analysis and other methods. The results found that the articles that received neutral PRC and negative PRC, or positive PRC and negative PRC, were no significant difference in the citations with the articles that received no post-publication PRC from the control group. The citations of articles that received positive PRC after publication were significantly higher than those that did not receive post-publication PRC.

**Analyzing the influence factors of the citations**
Citations have long been a critical indicator in evaluating academic influence. Scholars have made ongoing exploration on the reasons for being cited. Publishing papers in journals with high influence can improve the attention of research, obtain more readers, and then improve the citations (Bhandari *et al.*, 2007). Didegah *et al.* (2018) found that journal influence factors and international cooperation are two significant factors for the increase of the citations.

Many studies have pointed out a positive correlation between the number of references and the citations. The more references the paper has, the higher the citations will get (Webster *et al.*, 2009; Vieira and Gomes, 2010). Scholars Boyack and Klavans (2005) and Lancho-Barrantes *et al.* (2010) further found that the quality of references will also affect the citations. However, When Bornmann *et al.* (2012) studied the factors affecting the citations in the field of Chemistry (Angewandte Chemie International Edition, AE), they found that the citations are related to the journal language, the quality of references, the reputation of authors, and the research field. They did not find a significant relationship with the number of references.

In recent years, open peer review has attracted more and more scholars' attention. Some scholars found that the citations of papers in open access journals are higher than that in non-open access journals (Vanclay, 2013), and open peer review can improve the citations (Zong *et al.*, 2020). Ni *et al.* (2021) studied the published peer review opinions and the author's replies in Nature Communication from 2016 to 2017 and found that the opening of peer review opinions did not cause an increase in the citations, and the number of revised rounds of the manuscript was not significantly related to the citations. There was a weak negative correlation between the longer review comments and the citations. The reason for this may be related to the data used.

There are many fascinating studies in exploring influencing factors of the citations, such as more authors have more significant advantages in obtaining citation frequency (Chen, 2012), apers published at the beginning of the year will obtain more cited opportunities than at the end of the year (Ma *et al.*, 2019), review papers will get more citations than other types of papers (Bornmann, 2013), papers with short title will get more citations (Chawla, 2021; Haslam *et al.*, 2008),, papers supported by funds will get more citations than those do not obtained (Rohani *et al.*, 2021; Zhao, 2010), the number of formulas in the papers is negatively correlated with the citations (Annalingam et al., 2014), novelty and hot spot research will obtain more citations (Fu and Aliferis, 2010; Zhang *et al.*, 2021a,b; Roshani *et al.*, 2021).

To sum up, the recognition of academic article structures is attracting scholars' attention, and the use of deep learning technology to identify academic paper structures has achieved better results. In future research, it is necessary to use academic article structures information and give full play to the role of the structure information in academic articles mining-related research. The research on the mining of PRC has been paid more attention by scholars, mainly focusing on exploring the law of PRC, evaluating the quality of PRC, and analyzing PRC from the perspective of scientific measurement. By combing the research status of the analysis of the influencing factors of the citations, it can guide the follow-up research of this paper and explain the relevant findings of this paper. The open access to PRC and the rapid development of natural language processing and text mining technologies have led to the current research on the mining of PRC receiving greater attention. This paper combines PRC with academic articles and utilizes NLP and text mining technology to explore the distribution of PRC in different structures of academic articles.

## Methodology

**Research framework**

The research framework of this article is shown in Figure 1. The first step is data acquisition. We obtain the research data from the ACP journal website, convert the PDF document to plain text by the converter Xunjie [4], and obtain the corresponding bibliographic information of the articles from WOS, including title, author information, abstract, citation frequency and other information. Second, we use the feature words of the section title and the HAN model to identify the structural function of sections in ACP and extract the position information in PRC according to rules. Therefore, the PRC can be mapped to the original manuscript. We analyze the distribution of PRC in different structures of academic articles. The third step is to extract the feature words of PRC distributed in different structures by using the Chi-square testing and TF-IDF. Then, we analyze the distribution of feature words of PRC in different structures. Finally, the citations of articles are standardized by using the PCSI (paper citation standardized index) method. We explore whether the distribution of PRC in different structures is correlated to the citations. The methods we use include Spearman correlation coefficient, cumulative distribution function, K-S test and negative binomial regression analysis.

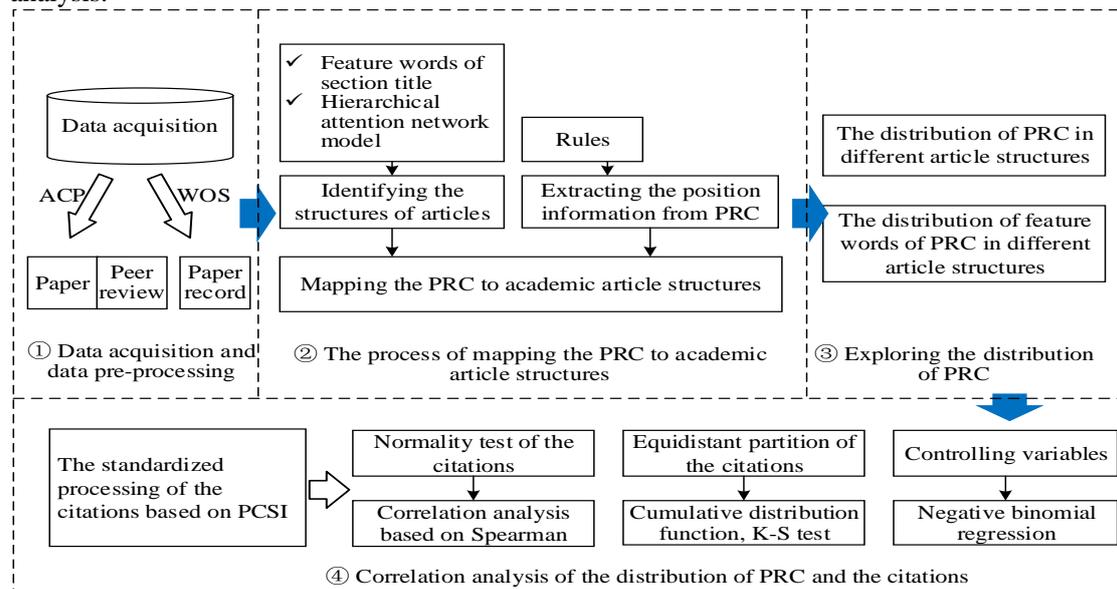

**Figure 1. The framework of exploring the distribution of PRC in different article structures**

## Data and methods

We introduce the experiment data and methods we used in this section. The main methods involve the hierarchical attention network model, rules matching, TF-IDF, CHI, topics clustering and the citations' standardizing.

### The introduction of data

Most journals do not publish the revision process of the manuscript, that is, the first manuscript the revised manuscript-the final accepted manuscript. The first manuscript is the direct evaluation object of the referees. Taking the first manuscript as research object is the premise of accurately analyzing the distribution of PRC in different structures. The data opened by ACP include the first manuscripts, all revised versions, final published manuscripts, peer review reports, author responses, downloads, view count, citations and other information. Therefore, this article takes the ACP journal as an example and uses the data of the first manuscript and peer review reports. The data we used contained 7,279 papers from 2001 to 2016, and the paper record such as title, abstract, author, institution, citations and other information come from the Web of Science (WOS).

### The methods used in this article

*Identifying the article structures based on feature words of section title*

The structures of scientific articles can usually be divided into introduction, materials & methods, results and discussion (IMRaD). Each structure represents different function, as detailed in Table 1. This article first annotates the structure of each section according to the feature words contained in the section title, and the feature words corresponding to the structure are shown in Table 2. The annotated data set is constructed for the machine learning and deep learning models. The HAN model will identify the sections whose title does not have feature words.

**Table 1. The illustration of the IMRaD**

| Structures | Illustration |
|---|---|
| Introduction | Explain the reasons or purpose for conducting the research, introduce the background of the research and introduce the relevant work |
| Materials & Methods | Introduce the research methods/design/materials/data utilized in the paper |
| Results | Analyze the results and illustrate the findings by using text/figures/ tables… |
| Discussion | Discuss the significance of the results, restate the main findings, articulate strengths and weaknesses and implications for other studies, state unresolved problems and future research to be undertaken |

**Table 2. The feature words of the section title corresponding to the IMRaD**

| Structure | Feature words |
|---|---|
| Introduction | introduction, motivation, background, overview, review of literature |
| Materials & Methods | system, theory, method, methods, methodology, model, models, framework, approach, approaches, methodologies, experimental, experiment, experiments, data, data and methods |
| Results | result, results, analysis |
| Discussion | discussion, discussions, conclusion, conclusions, summary, concluding, summary and conclusions |

*Identifying the article structures based on hierarchical attention network model*

The literature (Qin and Zhang, 2020) compared the performance of traditional machine learning models, deep learning models, Bert (Devlin *et al.*, 2018) model and HAN models with different granularities in the structure recognition of academic articles. It found that the sentence-level HAN model with the encoder Bi-LSTM+Attention was the best, and its structure diagram is shown in Figure 2.

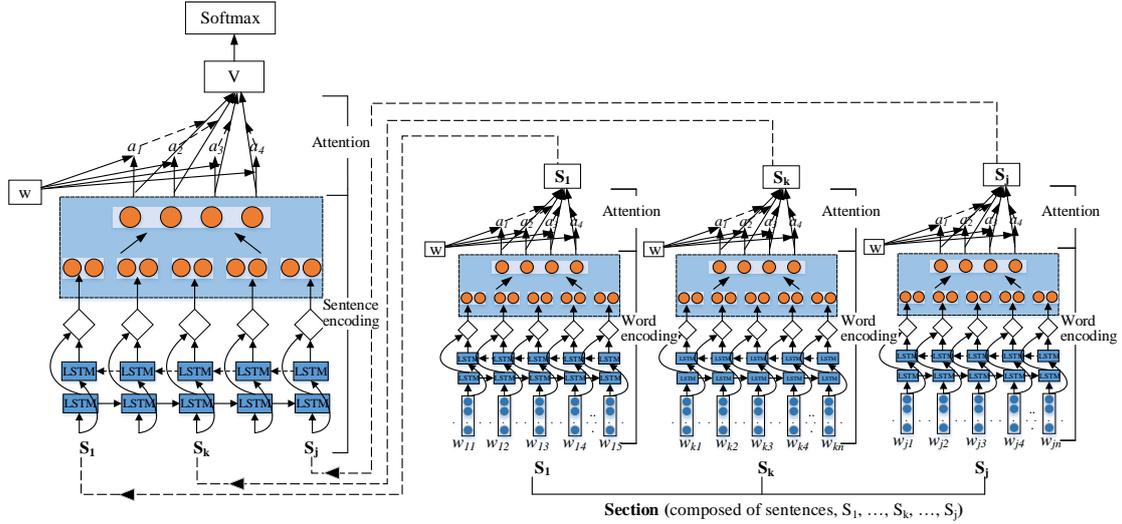
**Figure 2. The sentence-level Hierarchical Attention Network with encoder Bi-LSTM+Attention**

The sentence-level HAN model was proposed by Yang *et al.* (2017) to get better results in text classification. Starting from the objective phenomenon that sentences are composed of words and documents are composed of sentences, the model can encode words to represent sentences and encode sentences to represent documents, enabling the model to represent the documents from a sentence-word perspective. After each encoding, an attention layer is added to capture the information of words and sentences with higher weight. Specifically, in the process of identifying the structures of the academic articles, the sections are first split into sentences according to punctuations and using the encoder Bi-LSTM + Attention to encode the words in each sentence (Eq. (1)- Eq. (4)). The encoding results are input to the Attention layer, and taking the output of this layer as the original representation of sentences (Eq. (5)). Subsequently, using the same encoder to encode the sentences shown as Eq. (6)-Eq. (9). The encoding results are input to the Attention layer, and taking the output as the representation of sections (Eq. (10)-Eq. (11)). At last, the final represent are input to the Softmax layer.

**Word encoder:**

$$x_{ki} = W_e w_{ki}, k \in [1,j],\ i \in [0,n] \quad (1)$$

$$\vec{h}_{ki} = \overrightarrow{LSTM}(x_{ki}) \quad (2)$$

$$\overleftarrow{h}_{ki} = \overleftarrow{LSTM}(x_{ki}) \quad (3)$$

$$h_{ki} = (\vec{h}_{ki}, \overleftarrow{h}_{ki}) \quad (4)$$

**Word attention:**

$$a_{ki} = \frac{\exp(\tanh(W_w \cdot h_{ki} + b_w)^T \cdot W)}{\sum_{i=0}^{n} \exp(\tanh(W_w \cdot h_{ki} + b_w)^T \cdot W)} \quad (5)$$

**Sentence encoder:**

$$S_k = \sum_{i=0}^{n} a_{ki} h_{ki} \quad (6)$$

$$\vec{h}_k = \overrightarrow{LSTM}(S_k) \quad (7)$$

$$\overleftarrow{h}_k = \overleftarrow{LSTM}(S_k) \quad (8)$$

$$h_k = (\vec{h}_k, \overleftarrow{h}_k) \quad (9)$$

**Sentence attention:**

$$a_k = \frac{\exp(\tanh(W_w \cdot h_k + b_w)^T \cdot W)}{\sum_{i=0}^{n} \exp(\tanh(W_w \cdot h_k + b_w)^T \cdot W)} \quad (10)$$

$$V = \sum_{k=1}^{j} a_k h_k \quad (11)$$

In Eq. (1), $W_e$ is the embedding matrix for word vectors, $x_{ki}$ is the vector representation of word $w_{ki}$. In Eq. (6), the sentence $S_k$ is denoted as $(w_{k0}, w_{k1}, ..., w_{kn})$, which means that it consists of *n* words. $W_w$, $W$ and $b_w$ in Eq. (5) and Eq. (10) are randomly initialized weights and bias vector.

*Extracting the position information from the PRC based on rules*

It is necessary to extract the location information contained in PRC while exploring the distribution of PRC in different structures of academic articles. We divide the position information in PRC into two categories; one is the explicit position information, such as page, line, figure, table, equation, section and so on. The PRC can be mapped to the original manuscript according to the explicit position information. The corresponding structure of the position is the structure the PRC distributed. Examples of the rules we use are shown in Table 3, and all 46 rules can be obtained from the website of https://github.com/kakabular/peer-review (the rule of "filter" is used to filter the symbols that may be contained in the return value). While the PRC does not contain the explicit position information mentioned above, we take the feature words of section title (in Table 1) included in PRC as the implicit position information. For example, the feature words "model" in PRC represents that the PRC is distributed in the structure of Materials & Methods. For more accurately obtaining the corresponding structure of the PRC, the explicit position information is preferred to determine the distribution of the PRC. Taking a manuscript with six sections as an example, we identify the structure to which section belongs firstly, and map the PRC to the structure of the academic paper according to the position information, as shown in Figure 3. The section in Figure 3 corresponds to the first-level heading, and the number is the first-level heading number.

**Table 3. Examples of the rules for extracting the position information**

| Type | Rule | Example |
|---|---|---|
| page, line | @"page([\s\S]{1,10}?)line([\s\S]{1,5}?)[@–'""-/).,:a-z]" | page 10, line 7 |
| | @"p[.]([\s\S]{1,10}?)l[.n]([\s\S]{1,5}?)[@–'""-/).,:a-z]" | p.10, l.7 |
| | @"pg[.]([\s\S]{1,10}?)l[.n]([\s\S]{1,5}?)[@–'""-/).,:a-z]" | pg. 10, ln 7 |
| | @"pg([\s\S]{1,10}?)line([\s\S]{1,5}?)[@–'""-/).,:a-z]" | pg 10, line 7 |
| | @"p([\s\S]{1,10}?)/l([\s\S]{1,5}?)[@–'""-/).,:a-z]" | p 10/l 7 |
| equation | @"equation([\s\S]{1,5}?)[@–'""-/).,:a-z]" | equation (5) |
| | @"eq[.]([\s\S]{1,5}?)[@–'""-/).,:a-z]" | eq. 5, eq. (5) |
| table | @"table([\s\S]{1,5}?)[@–'""-/).,:a-z]" | table 5 |
| figure | @"fig[.]([\s\S]{1,5}?)[@–'""-/).,:a-z]" | fig.5 |
| | @"figure([\s\S]{1,5}?)[@–'""-/).,:a-z]" | figure 5 |
| section | @"section([\s\S]{1,3}?)[@–'""-/).,:a-z]"; | section 3.1 |
| | @"([1-8])[.][1-8]"; | 3.1 |
| filter | Replace(",", ""), Replace("–", ""), Replace(".", ""), Replace("s", ""), Replace("(", ""), Replace(" ", ""), Replace(":", "") Replace("–", "") | |

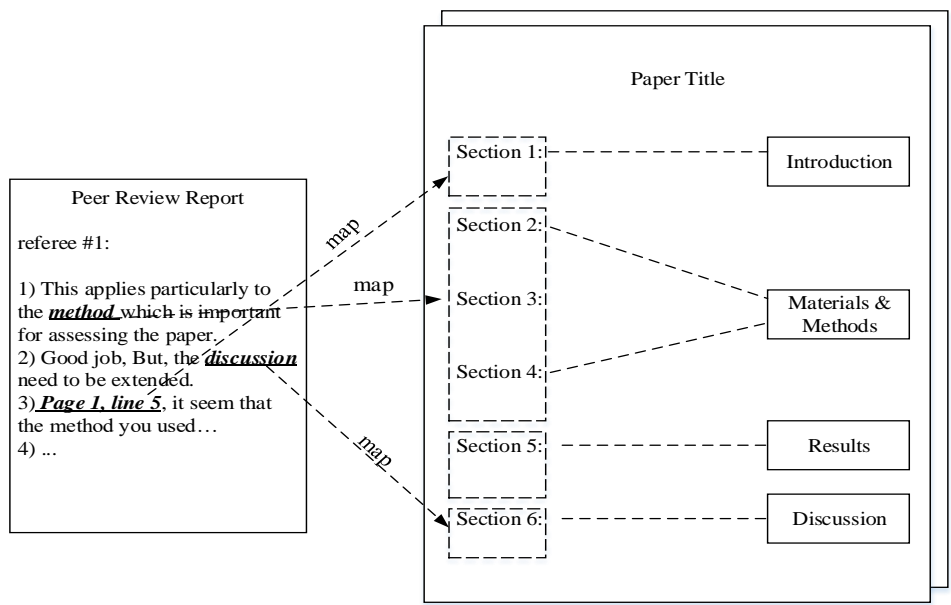

**Figure 3. The diagram of PRC mapped to the structures of academic article**

*Extracting the feature words of PRC distributed in different structures of academic articles*

First, it is necessary to preprocess the PRC, that is, to segment the PRC, filter the stop words and utilize the part of speech reduction tool to restore the words. After reduction, the words with complete meaning are more suitable for fine-grained text analysis than stemming. The tools used in preprocessing are the NLTK toolkits commonly used in natural language processing.

Subsequently, using the Chi-square test method to extract the feature words of PRC distributed in different structures. The Chi-square test assumes that the word $w$ is independently uncorrelated with the class $C_k$, and the difference between the observed value $x_i$ and the expected value E is calculated by Eq. (12). The greater the difference, the null hypothesis is not valid, which means that the word $w$ strongly correlates with the class $C_k$. Usually, the top *n* words with high Chi-square values are chosen as the feature words.

$$CHI\ (w, C_k) = \sum_{i=1}^{n} \frac{(x_i - E)^2}{E} \qquad (12)$$

The Chi-square test has a "defect of low-frequency word" problem. The returned results may contain words with low frequency but high Chi-square value. The TF-IDF method can extract words with significant differences in different classes, so this method is also used to extract feature words distributed in different structures in this paper. The returned results of the two methods (Chi-square test and TF-IDF) are sorted by value from high to low. Taking the intersection of the returned top 1,000 words of the result separately (setting this threshold is to cover as many feature words as possible) as candidate feature words, then excluding no-nouns and unseen character combinations manually to determine the final feature words.

*Standardizing the citations of academic articles*

In order to make the citations of papers published in different years and with different research topics comparable, it is necessary to standardize the citations. The process of standardizing the citations involves two parts: standardizing calculation method and research topics clustering.

- *The process of standardizing the citations based on PCSI*

Wu *et al.* (2020) considered that it is not feasible to compare the citations of two papers published in different disciplines and different years because of the differences in research fields and the time lag of the citations. They proposed a standardized method for eliminating the differences, named PCSI (paper citation standardized index). Several studies have shown significant differences in the citations with different research topics (Daud *et al.*, 2019; Park *et al.*, 2021). In this paper, we utilize the work of Wu *et al.* (2020) mentioned above, and the discipline is replaced with the research topic. The process of standardizing the citations of an article is as follows:

(1) If the citation is 0, the standardized citation is also 0;
(2) Set $y_t = \ln x_t$, and $z_t = (y_t - \bar{y}_t)/S_t$, where $\bar{y}_t$ represents the average of all papers published in the same topic and the same year with $e$ as the base of the logarithm, and $S_t$ represents the variance after the logarithm of the citations of all papers published in the same topic and the same year. $x_t$ is the citation of an article accumulated from a certain year to March 2022, which was published in a certain topic;
(3) The standardized citation of an article is $e^{z_t}$.

- *The method of research topics clustering based on hierarchical clustering ideas*

The method of clustering we used in this paper was proposed in the work of Qin *et al.* (2021). In this paper, the Doc2Vec model (Le and Mikolov, 2014) was used to generate vectors for abstracts of the experiment data. The Doc2Vec model can encode the long text (such as sentences and paragraphs) to fixed-length vectors. The generated vectors are represented as $W_{abstract_i}(v_1, v_2, v_3, \ldots, v_n)$, and we use cosine to calculate the similarity between abstracts denoted as eq. (13).

$$Sim(W_{abstractt_i}, W_{abstract_j}) = \frac{\sum_{i,j=1}^{n}(W_{abstract_i} \times W_{abstract_j})}{\sqrt{\sum_{i=1}^{n} W_{abstract_i}^2} \times \sqrt{\sum_{j=1}^{n} W_{abstract_j}^2}} \qquad (13)$$

The clustering process consists of two phases: the allocation phase and the transfer phase. Cluster quality is assessed by the DBI index. The smaller the DBI value, the better the clustering results.

*Allocation phase*

Select one of the abstracts vectors as the center of the initial class. In this phase, the algorithm follows a three-step procedure:
1. Traverse the elements in the set of abstracts except for the selected element.
2. Calculate the similarity of the selected vectors and the existing classes.
3. Add the abstract to the cluster with the largest similarity if the similarity reaches the minimum threshold; otherwise, a new cluster is created, and the abstract is used as the center of the new cluster.

*Transfer phase*

Multiple clusters are generated during the allocation phase, but abstracts in a cluster may have a greater similarity value than a certain cluster generated later. Therefore, it is necessary to compute the similarity of the current cluster and the clusters generated later. If the elements of the current cluster have higher similarity with a certain center of clusters generated later, add the elements to the cluster and delete them from the raw cluster.

*Correlation analysis methods*

In this paper, four methods are used to analyze the correlation between the citations and the distribution of PRC in different structures of academic articles to make empirical results scientific and reliable, including the Spearman, the cumulative distribution function, the K-S test and the negative binomial egression. In the negative binomial regression analysis, six variables, including paper type, title length, paper length, nationality, institute, author count and international cooperation, are controlled.

# Results

This section introduces the results of the identification of the structures of academic articles, the results of the extraction of position information in PRC, and the construction of experimental data sets; subsequently, the distribution of PRC in different structures, the distribution of the feature words of PRC in different structures, and the distribution of PRC in different structures while considering the citations are analyzed.

**The results of identifying the results of structures of academic articles**

The structures of the ACP journal papers are first identified according to the feature words of section title, and the results are shown in Table 4. "Others" indicates the sections whose structure cannot be identified according to the feature words, and the number is 7,688 distributed in 3,767 papers. Two thousand seven hundred forty-three (2,743) pieces of data in each structure are selected to build a balanced data set, and the data set is divided into 8:2 (training set and test set) in the traditional machine learning models and the data set is divided into 6:2:2 (training set, validation set and test set) in the deep learning models.

Table 4. The results of identifying the structures of articles by feature words of section title

| Journal | Time | Paper count | Introduction count | Materials & Methods count | Results count | Discussion count | Others count |
|---|---|---|---|---|---|---|---|
| ACP | 2001-2016 | 7,279 | 7,067 | 7,243 | 2,743 | 7,397 | 7,688 |

The identification results of article structures by using the traditional machine learning model XGB, BERT model, basic neural network model, sentence-level HAN model and paragraph-level HAN model in the ACP data are shown in Table 5. From Table 5, it can find that the sentence-level HAN with Bi-LSTM+Attention as the encoder has the best performance, which is better than the traditional machine learning model, the BERT model, the basic neural network model and the paragraph-level HAN. The Macro-F1 value of sentence-level HAN is 0.8661. Therefore, this article utilizes the sentence-level HAN with the encoder Bi-LSTM+Attention to identify the structure to which the "Others" section belongs.

Table 5. The results of identifying the structures of articles by models

| Type | Model | | Macro-P | Macro-R | Macro-$F_1$ | Accuracy-I | Accuracy-M | Accuracy-R | Accuracy-D |
|---|---|---|---|---|---|---|---|---|---|
| Baseline | XGB | TF | 0.8262 | 0.8277 | 0.8262 | 0.8721 | 0.7592 | 0.8716 | 0.8081 |
| | | CHI | 0.8316 | 0.8324 | 0.8317 | 0.8721 | 0.7900 | 0.8613 | 0.8061 |
| | | IG | 0.8224 | 0.8234 | 0.8226 | 0.8637 | 0.7726 | 0.8530 | 0.8042 |
| | BERT Model | | 0.8480 | 0.8491 | 0.8482 | 0.9368 | 0.7684 | 0.8125 | 0.8787 |
| Basic-NN | Bi-LSTM+CNN | | 0.8260 | 0.8294 | 0.8266 | 0.8580 | 0.7988 | 0.7663 | 0.8809 |
| | Bi-LSTM+Attention | | 0.8513 | 0.8525 | 0.8511 | 0.9177 | 0.7800 | 0.8685 | 0.8390 |
| | Bi-LSTM+CNN+Attention | | 0.8362 | 0.8365 | 0.8362 | 0.8802 | 0.8129 | 0.8081 | 0.8438 |
| Sentence-level HAN | Bi-LSTM+CNN | | 0.8137 | 0.7985 | 0.7885 | 0.9614 | 0.8367 | 0.5068 | 0.8891 |
| | Bi-LSTM+Attention | | 0.8668 | 0.8666 | 0.8661 | 0.9672 | 0.8186 | 0.8004 | 0.8801 |
| | Bi-LSTM+CNN+Attention | | 0.8533 | 0.8498 | 0.8488 | 0.9537 | 0.8745 | 0.7117 | 0.8592 |
| Paragraph-level HAN | Bi-LSTM+CNN | | 0.8408 | 0.8398 | 0.8396 | 0.9196 | 0.8253 | 0.8113 | 0.8032 |
| | Bi-LSTM+Attention | | 0.8616 | 0.8601 | 0.8604 | 0.9123 | 0.8246 | 0.8519 | 0.8515 |
| | Bi-LSTM+CNN+Attention | | 0.7866 | 0.7472 | 0.7417 | 0.6561 | 0.5293 | 0.8842 | 0.9194 |

**The results of extracting the position information in PRC**

The position information in PRC is extracted through the rules described in the above section. The count distribution of PRC contained position information over the years is shown in Figure 4. The coverage rate in the legend is the ratio of the PRC count with position information to the number of all PRC in the same year. It can find that the coverage rate is from 0.77 to 0.83 in Figure 4. The reason for this is that there are significant differences in the style of PRC writing, and there are indeed some PRC that do not contain position information.

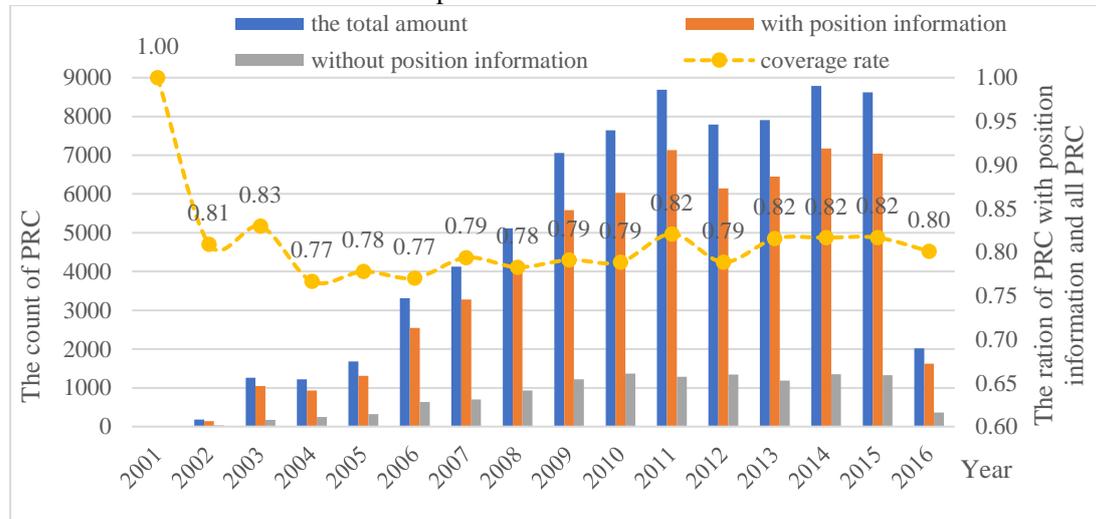

Figure 4. The distribution of PRC with position information in different years.

**The construction of experimental data set for the distribution of PRC**

After identifying the structures of the academic articles, we construct two datasets for analyzing the distribution of PRC in different structures. The articles with the completed IMRaD structures are selected as experimental data, consisting of 2,687 papers and 5,797 peer review reports. Specifically, structures of 1,333 articles are identified by the feature words of the section title, denoted as dataset-A; structures of 1,354 papers are identified by the feature words of the section title and the sentence-level HAN, denoted as dataset-B. The distribution of the two constructed datasets from 2001 to 2016 is shown in Figure 5.

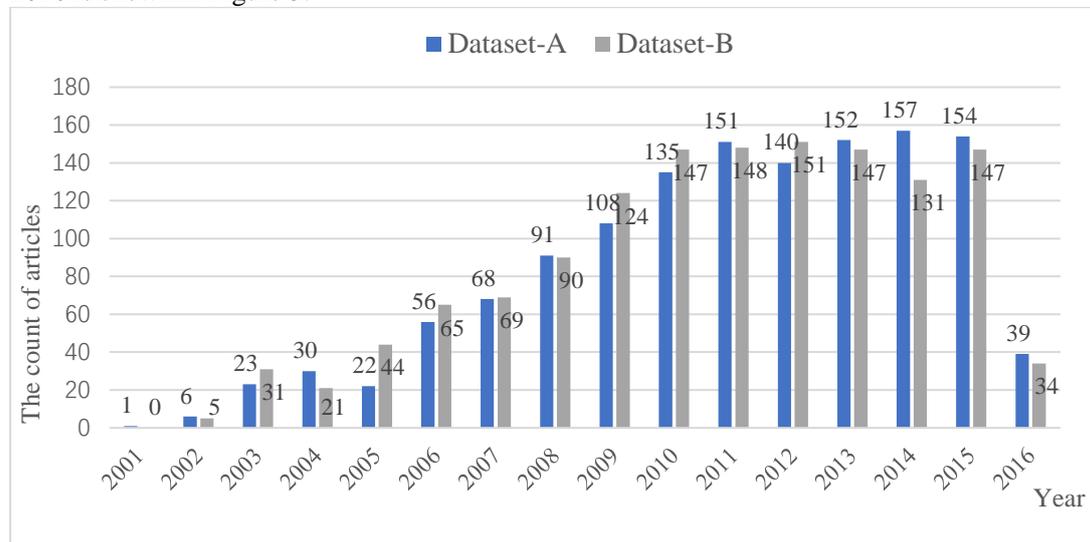

Figure 5. The distribution of constructed datasets from 2001 to 2016.

The purpose of dividing experimental data into two datasets was to analyze whether there were differences in the distribution of PRC in different structures identified fully by the feature words of section title and those identified by the feature words and sentence-level HAN model. If there is a significant difference, it is explained that the accuracy of identifying the structures of academic articles by the sentence-level HAN model needs to be improved. Otherwise, the HAN model with the encoder Bi-LSTM+Attention is verified from the side to have a better effect on the recognition of the structures.

## The distribution of PRC in different article structures

For answering the RQ1, this section analyzes the distribution of PRC in different structures of academic papers, taking the data of 2009 as an example shown in Figure 6.

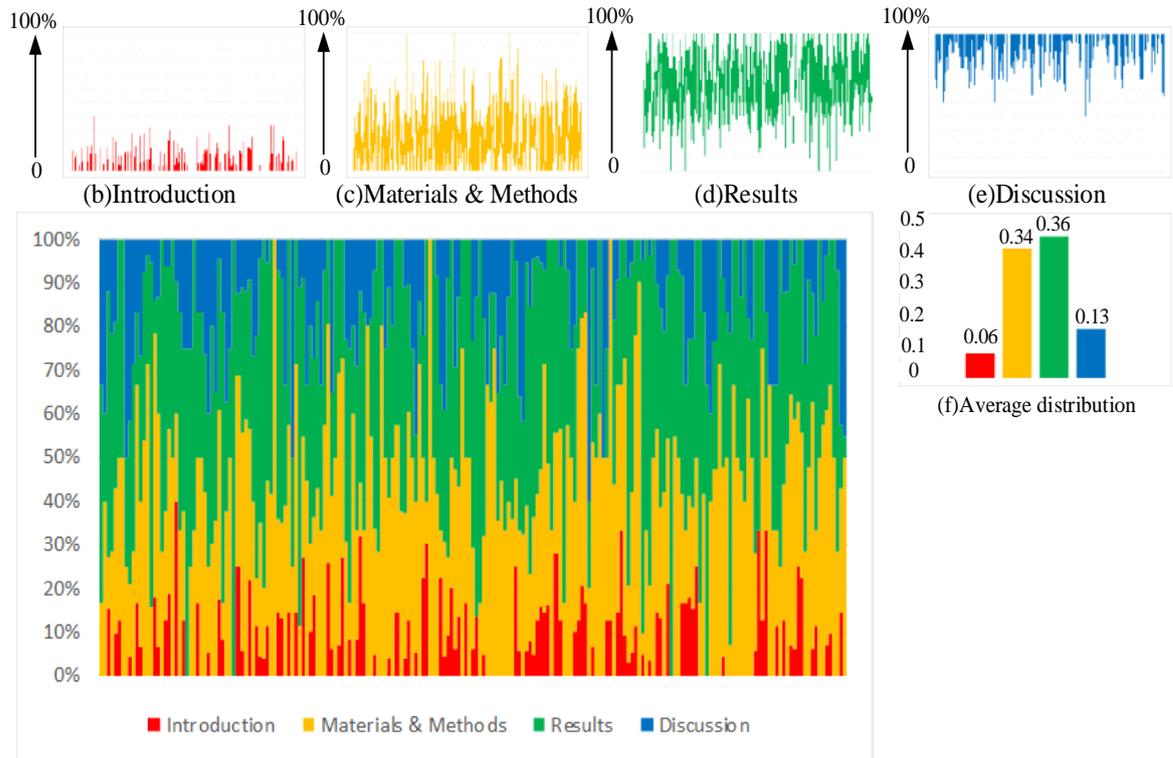

(a)The distribution of PRC in IMRaD.

**(1)Dataset-A (2009)**

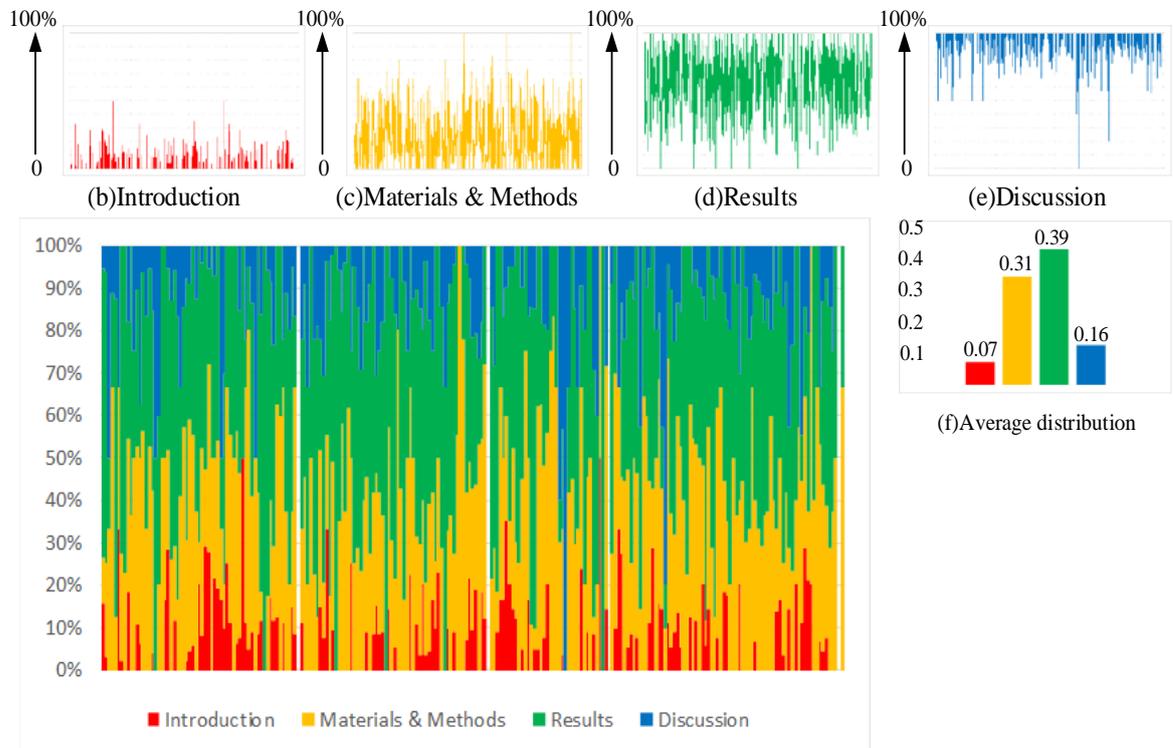

(a)The distribution of PRC in IMRaD

**(2)Dataset-B (2009)**

**Figure 6. The distribution of PRC in IMRaD of articles in 2009.**

Figure 6(1a) and Figure 6(2b) represent the overall distribution of PRC in different structures of academic papers in dataset-A and dataset-B, respectively. In most cases, the proportion of PRC distributed in Materials & Methods and Results is significantly higher than that of Introduction and Discussion, reflecting that materials, data, methods and experimental results are more focused on by referees.

The decomposed results of Figure 6(1a) and Figure 6(2a) are shown in Figure 6(1)(b)-Figure 6(1e) and Figure 6(2b)-Figure 6(2e), which can be intuitively found that there are more PRC distributed in the structures of Materials & Methods and Results than the structures of Introduction and Discussion.

Figure 6(1f) and Figure 6(2f) show the average distribution of PRC in different structures. The average distribution of PRC in Materials & Methods and Results is significantly higher than that in the structures of Introduction and Discussion.

Also, we can find that the sentence-level HAN with Bi-LSTM+Attention as the encoder has a good performance in identifying the structures of academic articles from the similar distribution of the subgraphs of Figure 6(1) and Figure 6(2).

The distribution of PRC in different structures each year is different. We use the average distribution of PRC over the years to analyze this question. (Note: the distribution of PRC in different structures in 2003-2016 can be found in Figure A1 and A2 of Appendix A, respectively).

Figure 7 and Figure 8 shows the average distribution of PRC in different structures in dataset-A (2001-2016) and dataset-B (2002-2016). It can be found that the average distribution of PRC in the four structures each year is relatively stable, and the PRC distributed in the structures of Materials & Methods and Results is significantly higher than that in the structures of Introduction and Discussion, indicating that the referees typically pay more attention to Materials & Methods and Results. In addition, we find that referees always pay less attention to the structure of the Introduction than the other three structures in most years.

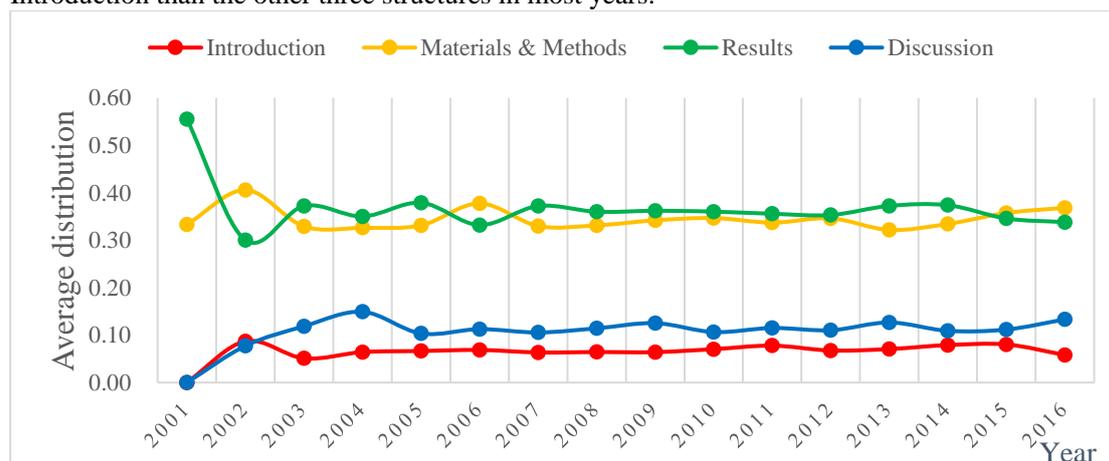

**Figure 7. Dataset-A: The average distribution of PRC in IMRaD**

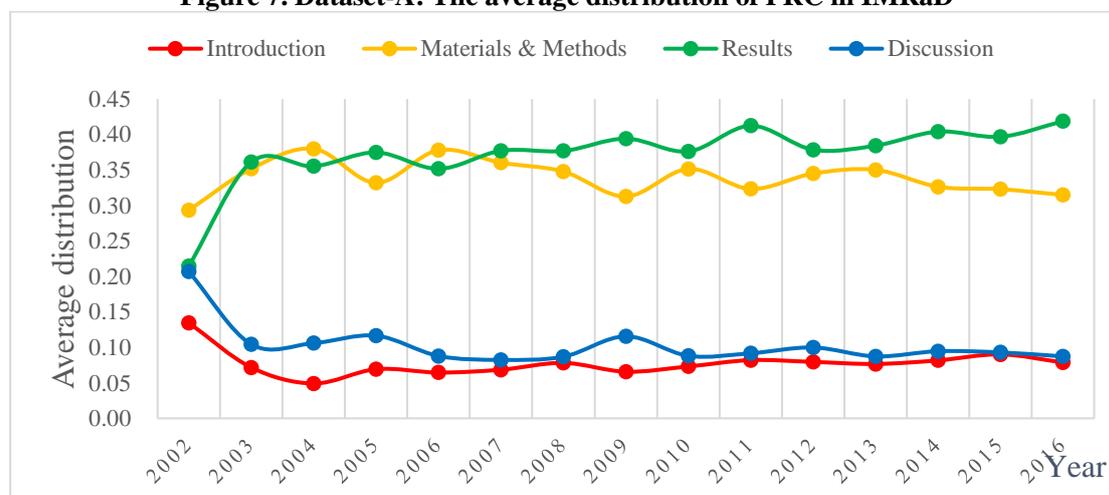

**Figure 8. Dataset-B: The average distribution of PRC in IMRaD**

### The distribution of feature words of PRC in different article structures

The distribution of PRC in different structures of articles is analyzed in the above section, and the macro conclusion is that the structures of Materials & Methods and Results are more concerned by referees than the other two structures. We continue to explore what specific content referees focus on in different structures.

For answering the RQ2, this section uses the feature word extraction methods mentioned above. The return results of the two methods are sorted by the CHI value and the TF-IDF value from high to low. Both take the top 1,000 words and find the intersection of the return results as candidate feature words. The number of candidate feature words distributed in each structure is 142, 401, 397 and 175 (sorted by IMRaD structure). After manual filtering, the number of feature words distributed in each structure is 60, 119, 137 and 83 (the order is the same as above). The distribution of the feature words of PRC in different structures reflects the details that the referees focus on. This section selects the top 60 feature words from each structure for visualization shown in Figure 9 and analyzes the distribution of feature words in different structures.

In Figure 9, different feature words are represented by different colors, and the height of the color block represents the ratio of the frequency of the feature word mentioned in the structure to the total number of PRC distributed in the corresponding structure. So, the importance of the feature words in different structures can be measured.

The Introduction structure typically describes the research background, reviews the relevant research, and briefly explains the research objectives, research content, research methods, and research significance. From the distribution of the feature words in the Introduction of Figure 9, we can find that "introduction", "author", "reference", "background" and other feature words have a larger proportion. Taking the word "introduction" as an example, although it is mentioned in all four structures, the highest proportion is in the structure of Introduction, reaching 0.1638, and the proportion of the remaining three structures is 0.0096 (Materials & Methods), 0.006 (Results) and 0.0144 (Discussion).

The structure of Materials & Methods usually details the materials/data/methods used in articles. In Materials & Methods of Figure 9, feature words such as "model", "data", "method", "measurement", "parameter", "experiment", "approach", "system" occupy a higher proportion. Taking the word "data" as an example, it is mentioned in all four structures, but the highest proportion is in Materials & Methods, with a value of 0.2126. The other three are 0.0651 (Introduction), 0.1116 (Result) and 0.0948 (Discussion).

The Results structure analyzes and illustrates research findings using text, figures and tables. From the distribution of feature words in Result of Figure 9, we can find that "figure", "result", "measurement", "data", "analysis", "values" and other feature words have a higher proportion. Taking the word "figure" as an example, the highest proportion is in Results, with a value of 0.4326. The remaining three structures have values of 0.0864 (Introduction), 0.1201 (Materials & Methods) and 0.1661 (Discussion).

The Discussion structure is to discuss the significance of the research, restate the main findings and conclusions, clarify the strengths and weaknesses in the research, explain the implications for future work and describe the following work plan. From the distribution of feature words in Discussion of Figure 9, we find that the feature words such as "discussion", "conclusion", "figure", "result", and "data" have a higher proportion. Taking the word "discussion" as an example, the highest proportion is in Discussion, with a value of 0.2963. The proportion of the remaining three structures is 0.0391 (Introduction), 0.0355 (Materials & Methods) and 0.0538 (Results).

The above analysis shows that the distribution of feature words of PRC in different structures is closely related to the structure, reflecting the detailed content that the referees pay attention to in different structures.

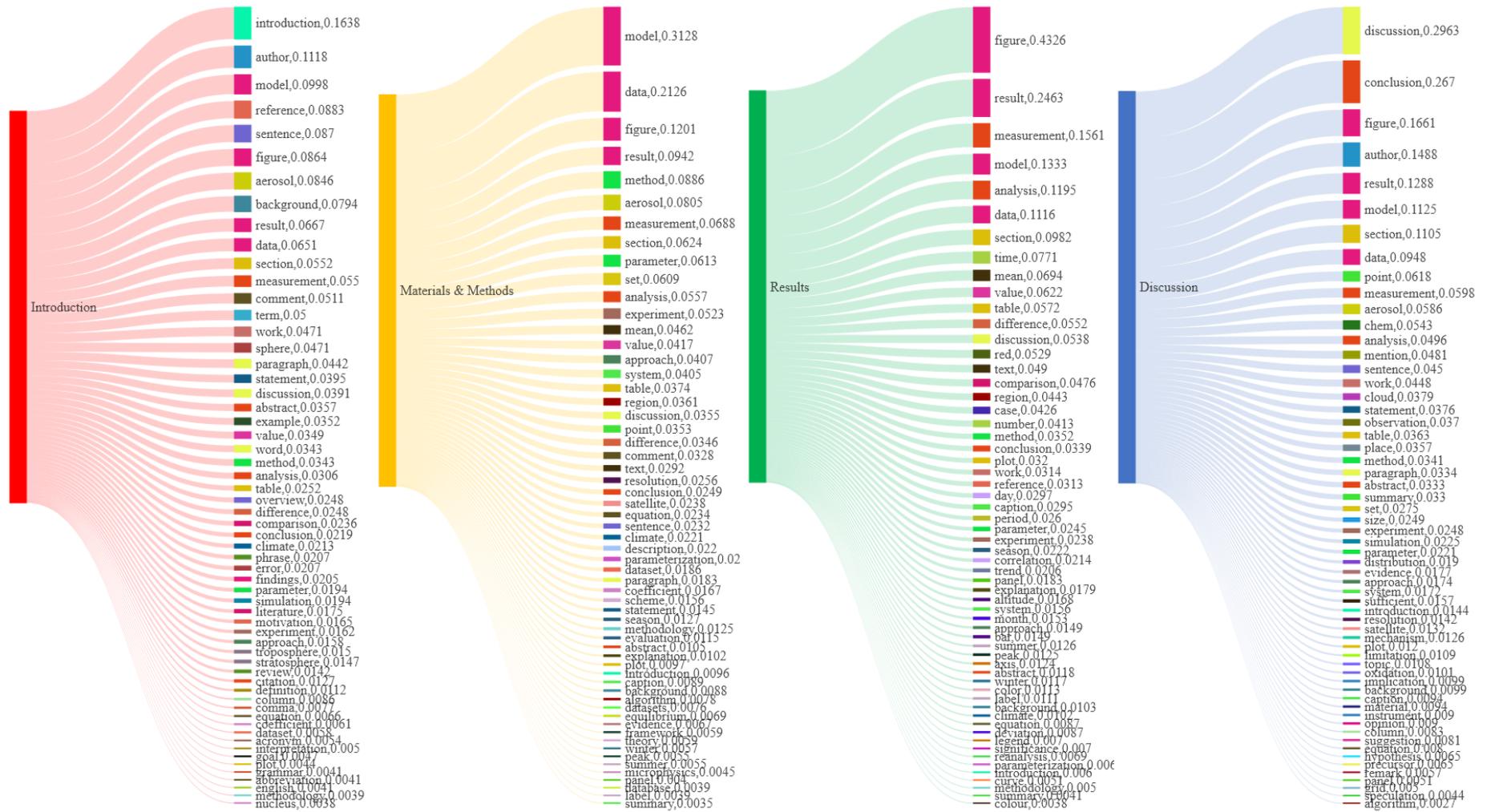

Figure 9. The distribution of feature words of PRC in IMRaD

**Correlation analysis of the distribution of PRC in different structures of academic articles and the citations of academic articles**

For answering the RQ3, this section uses four methods, namely spearman correlation coefficient, cumulative distribution function, K-S test and negative binomial regression, to analyze whether the distribution of PRC in different structures is related to the citations of academic articles. The purpose of using four methods is to ensure that the conclusions are reliable. The proportion of PRC in the different structures is denoted as *I, M, R* and *D*. Due to the small number of papers published in 2001-2002, the citations published after 2016 have not yet stabilized, so these papers and corresponding review reports are not considered. Therefore, the number of papers involved in this section is 2,503 from 2003 to 2015, and the number of peer review reports is 5,419.

*Correlation analyzing by using the Spearman correlation coefficient*

We first conduct a normal test of the standardized citations by using Q-Q plot and K-S test, and the results can be found in Appendix B. Since the citations do not conform to the normal distribution, we use Spearman to analyze the correlation between the distribution of PRC in different structures of academic papers and the citations. The correlation coefficients are shown in Table 6. From Table 6, it can be found that there is no significant correlation between the distribution of PRC in different structures and the citations.

**Table 6. Correlation analysis based on the Spearman correlation coefficient**

|  | I | M | R | D |
|---|---|---|---|---|
| **Citations** | -0.011 (0.590) | -0.010 (0.621) | 0.001 (0.965) | 0.028 (0.166) |

Note：*$p<0.05$, **$p<0.01$

*Correlation analyzing by using the cumulative distribution function and K-S test*

In this section, the experimental data are sorted in descending order of the citations. The equidistant partition is conducted at 10% intervals, and 10 partitions are obtained. The information of paper count, maximum citations, minimum citations and average citations in each partition is shown in Table 7.

**Table 7. The information of ten partitions**

| Partition | Paper count | Max. | Min. | Ave. | Partition | Paper count | Max. | Min. | Ave. |
|---|---|---|---|---|---|---|---|---|---|
| Top 10% | 251 | 14.3017 | 3.5145 | 5.2431 | Top 50-60% | 251 | 0.9971 | 0.7635 | 0.8711 |
| Top 10-20% | 251 | 3.507 | 2.3674 | 2.8523 | Top 60-70% | 251 | 0.7596 | 0.5433 | 0.6468 |
| Top 20-30% | 251 | 2.3553 | 1.7609 | 2.0299 | Top 70-80% | 251 | 0.5425 | 0.3869 | 0.4645 |
| Top 30-40% | 251 | 1.7575 | 1.3035 | 1.5053 | Top 80-90% | 251 | 0.3863 | 0.2592 | 0.3238 |
| Top 40-50% | 251 | 1.3025 | 0.9973 | 1.1451 | Top 90-100% | 244 | 0.259 | 0.0502 | 0.1858 |

Selecting the top 10%, top 20-30%, top 50-60%, top 70-80% and top 80-90% five partitions indicate different citations (Note: The number of papers in the Top 90%-100% partition is different from other partitions, so the Top 80-90% is used instead of Top 90-100%), and the cumulative distribution of PRC in different structures and different partitions is shown in Figure A4 of Appendix C.

Subsequently, The K-S test is used to calculate the similarity of the cumulative distribution of PRC in different citation partitions. We take the partition of the top 10% and top 80%-top 90% as an example, and the cumulative distribution similarity of PRC in different partitions is shown in Table 8. From Table 8, it can be found that the *p*-value of the K-S test results is greater than 0.05 in the structure of Introduction, Materials & Methods and Discussion, which means that the cumulative distribution of PRC has the same distribution in the highly cited partition and low cited partition. Therefore, the distribution of PRC in Introduction, Materials & Methods and Discussion is not related to the citations.

Table 8 also shows a significant difference in the structure of Results. Further analysis of the cumulative distribution similarity of PRC in different Partitions, we find that the p-value of K-S (Top 10%, Top 20-30%) is 0.0034, while the p-value of K-S (Top10%, Top 40-50%) is 0.4691. If the distribution of PRC in the structure of Results is related to the citations, the differences should be consistent with the increase of the cited partitions rather than contradictory. Therefore, the distribution of PRC in Results is independent of citations.

**Table 8. The p-value of K-S test between the partitions of top 10% and top 80-90%**

|  | I | M | R | D |
|---|---|---|---|---|
| K-S (Top 10%, Top 80%-90%) | 0.9888 | 0.4691 | 0.0434 | 0.2429 |

Note：*$p<0.05$, **$p<0.01$

*Correlation analyzing by using the negative binomial regression*
This section takes paper type, title length, author count, page count, reference count and international cooperation (determined according to the number of nationalities contained in the paper) as control variables and constructs a basic model. Four independent variables, namely the distribution proportion of PRC in Introduction, the distribution proportion of PRC in Materials & Methods, the distribution proportion of PRC in Results and the distribution proportion of PRC in Discussion, are added to the basic model in turn. Only one independent variable is added at a time. Descriptive information for all variables is shown in Table 9. Before performing regression analysis, a multicollinearity test is performed on all variables, and the VIF (variance inflation factor) value is less than 10.

**Table 9. Descriptive information of variables**

| Variable ID | Variable name | Variable attributes | | Maximum | Minimum | Average |
|---|---|---|---|---|---|---|
| X1 | Paper type | Review Others | Control variable | - | - | - |
| X2 | Title length | | Control variable | 38 | 4 | 14.6093 |
| X3 | Author count | | Control variable | 58 | 1 | 7.0967 |
| X4 | Page count | | Control variable | 42 | 4 | 15.9385 |
| X5 | Reference count | | Control variable | 225 | 4 | 56.3979 |
| X6 | International cooperation | Yes No | Control variable | - | - | - |
| X7 | Distribution proportion of PRC in I | | Independent variable | 0.3485 | 0.0000 | 0.0419 |
| X8 | Distribution proportion of PRC in M | | Independent variable | 0.5231 | 0.0000 | 0.1932 |
| X9 | Distribution proportion of PRC in R | | Independent variable | 0.5379 | 0.0000 | 0.2134 |
| X10 | Distribution proportion of PRC in D | | Independent variable | 0.3674 | 0.0000 | 0.0592 |

As shown in Table 10, there is no significant change in the regression coefficients and significance levels of the six control variables in the model M1-M5. Taking the basic model M1 as an example, the regression coefficient of the paper type is 0.099, and the significance level is 0.717, which is greater than 0.05. So, in this study, the paper type had no effect on the citations. Usually, review papers can obtain more citations, but the conclusion of this paper is the opposite. The possible reason is that the proportion of review papers in the experimental data is relatively small. ACP journal has published 126 review papers so far, and this experimental data only contains 20 papers. The regression coefficient of the title length is -0.012, and the significance level is 0.044, less than 0.05. So, the title length has a significant negative effect on the citations. The longer the title, the lower the citations, which is more consistent with the conclusions of existing studies, and the reason may be that shorter paper titles are easy to remember and retrieve. The author count has a regression coefficient of 0.018 and a significance level of 0.000, less than 0.05. Therefore, the author count has a significant positive effect on the citations, and when the author count is larger, a higher citation will be obtained. The possible reason is that more authors expand the visibility of papers and can increase the probability of self-citation. The regression coefficient of the page count is 0.015, and the significance level is 0.023, less than 0.05, which has a positive effect on the citations. The regression coefficient of reference count is 0.004, and the significance level is 0.002, less than 0.05. Therefore, the reference count has a significant positive impact on the citations. That means more references will obtain a higher citation. The possible reason is that more references show that the relevant research is more fully sorted out, compared and discussed, and the conclusions are rigorous and reliable. The regression coefficient of international cooperation is 0.057, and the significance level is 0.310, greater than 0.05. It indicates that international cooperation has no significant effect on the frequency of citations. The possible reason is related to the experimental data used in this paper, research on ACP usually involves several regions, and international cooperation is a common phenomenon. There are 1,320 papers with countries more than two in our experimental data.

**Table 10. The results of correlation analyzing by using the Negative Binomial Regression**

| Variable name | Regression coefficient | | | | |
|---|---|---|---|---|---|
| | Model M1 | Model M2 | Model M3 | Model M4 | Model M5 |
| Paper type | 0.099 (0.717) | 0.091 (0.738) | 0.100 (0.713) | 0.100 (0.714) | 0.084 (0.758) |
| Title length | -0.012* (0.044) | -0.012* (0.042) | -0.012* (0.044) | -0.011* (0.045) | -0.011* (0.048) |
| Author count | 0.018** (0.000) | 0.019** (0.000) | 0.018** (0.000) | 0.019** (0.000) | 0.018** (0.000) |
| Page count | 0.015* (0.023) | 0.015* (0.026) | 0.015* (0.023) | 0.015* (0.023) | 0.015* (0.020) |
| Reference count | 0.004** (0.002) | 0.004** (0.001) | 0.004** (0.002) | 0.004** (0.002) | 0.004** (0.002) |
| International cooperation | 0.057 (0.310) | 0.057 (0.311) | 0.057 (0.312) | 0.058 (0.300) | 0.059 (0.295) |
| Distribution proportion of PRC in I | | -0.444 (0.452) | | | |
| Distribution proportion of PRC in M | | | -0.031 (0.906) | | |
| Distribution proportion of PRC in R | | | | -0.133 (0.588) | |
| Distribution proportion of PRC in D | | | | | 0.619 (0.177) |
| Sample count | 2503 | 2503 | 2503 | 2503 | 2503 |
| Discrete test O-value | 21.249 | 21.249 | 21.249 | 21.249 | 21.249 |
| Discrete test P-value | 0.000 | 0.000 | 0.000 | 0.000 | 0.000 |
| AIC | 8412.850 | 8414.297 | 8414.836 | 8414.564 | 8412.956 |
| BIC | -18426.832 | -18419.559 | -18419.020 | -18419.293 | -18420.900 |
| $p$ | 0.000 | 0.000 | 0.000 | 0.000 | 0.000 |
| McFadden $R^2$ | 0.012 | 0.012 | 0.012 | 0.012 | 0.012 |

Note: *$p<0.05$, **$p<0.01$

In the model M2, the regression coefficient of the distribution proportion of PRC in I is -0.444, and the significance level is 0.452, which is greater than 0.05. So, the distribution proportion of PRC in I has no significant effect on the citations. Moreover, the value of AIC (Akaike Information Criterion) and BIC (Bayesian Information Criterion) are all increased compared with the basic model M1, indicating that the model is not well constructed with the addition of independent variables. In the model M3, the regression coefficient of the distribution proportion of PRC in M is -0.031 and the significance level is 0.906, which is greater than 0.05. So, the distribution proportion of PRC in M has no significant effect on the citations. Moreover, the value of AIC and BIC are all increased compared with the basic model M1, indicating that the model is not well constructed with the addition of independent variables. In the model M4, the regression coefficient of the distribution proportion of PRC in R is -0.133, and the significance level is 0.588, which is greater than 0.05. So, the distribution proportion of PRC in R has no significant effect on the citations. Moreover, the value of AIC and BIC are all increased compared with the basic model M1, indicating that the model is not well constructed with the addition of independent variables. In the model M5, the regression coefficient of the distribution proportion of PRC in D is 0.619, and the significance level is 0.177, which is greater than 0.05. So, the distribution proportion of PRC in D has no significant effect on the citations. Also, the value of AIC and BIC are all increased compared with the basic model M1, indicating that the model is not well constructed with the addition of independent variables.

In this section, we use four methods to analyze whether the distribution of PRC in different structures of academic articles is related to the citations. The above analysis shows that the results obtained by the four methods are consistent, proving that the conclusion is reliable. The conclusion is that the distribution of PRC in different structures is unrelated to the citations.

## Discussion

We obtain several findings after exploring the three research questions mentioned in Introduction. We discuss the findings, implications and limitation in this part

For answering the RQ1, we first utilize feature words of section title and sentence-level HAN model to identify the structure of academic papers. Then, we analyze the distribution of PRC in different structures. We find that referees typically pay more attention to Materials & Methods and Results than Introduction and Discussion in the constructed two datasets (dataset-A is determined entirely by the feature words of section title, and Dataset-B is determined by the feature words of section title and sentence-level HAN model). It indicates that the methods and results in articles are the focus of referees, and it also shows that the sentence-level HAN model has certain reliability. The recognition result of the model is not very ideal. Since the IMRaD structure has many advantages, publishing groups/journals should encourage authors to use the standardized structure when writing papers. In this way, the structural information will play an essential role in the research of full-text mining of academic articles. Moreover, researchers should ensure the scientificity of methods and the reliability of the results to obtain more publishing opportunities and improve the efficiency of scholarly communication.

For answering the RQ2, we combine TF-IDF with CHI to extract the feature words of PRC distributed in different structures. There are obvious differences in the distribution of feature words of PRC in each structure. Even if the same feature word appears in different structures, the weight is quite different, such as "model", "data", "figure", etc. The top 10 feature words with the highest weight in each structure compared with the other three structures shown in table 11. It should be noted that several feature words are from the research content, such as aerosol, age, chem, emission, etc., which is not well filtered by the method we use. In the structure of Introduction, authors should pay more attention to the literatures they cited. In the structure of Materials & Methods, model, data, method, parameter, etc. are the focus of referees. In the structure of Results, figure, result, measurement, mode, analysis, value, table, etc. are the focus of referees. In the structure of Discussion, discussion, conclusion and findings are the focus of referees. Therefore, the feature words of PRC distributed in different structures can reflect the specific content of referees' concern. Authors should take the writing very seriously to ensure the scientific quality of academic articles.

**Table 11 the top 10 feature words of PRC in each structure compared with the other three.**

| Introduction | | | | | Materials & Methods | | | | |
|---|---|---|---|---|---|---|---|---|---|
| Feature word | **I** | M | R | D | Feature word | I | **M** | R | D |
| introduction | 0.1638 | 0.0096 | 0.006 | 0.0144 | model | 0.0998 | 0.3128 | 0.1333 | 0.1125 |
| reference | 0.0883 | None | None | None | data | 0.0651 | 0.2126 | 0.1116 | 0.0948 |
| sentence | 0.0870 | None | None | None | age | None | 0.1712 | None | None |
| aerosol | 0.0846 | 0.0805 | 0.0657 | 0.0586 | method | 0.0343 | 0.0886 | 0.0352 | 0.0341 |
| background | 0.0794 | 0.0088 | 0.0103 | 0.0099 | chem | None | 0.0773 | None | None |
| comment | 0.0511 | None | None | None | parameter | 0.0194 | 0.0613 | 0.0245 | 0.0221 |
| term | 0.0500 | None | None | None | set | None | 0.0609 | None | None |
| work | 0.0471 | None | None | None | emission | None | 0.0554 | None | None |
| sphere | 0.0471 | None | None | None | observation | None | 0.0534 | None | None |
| Results | | | | | Discussion | | | | |
| Feature word | I | M | **R** | D | Feature word | I | M | R | **D** |
| figure | 0.0864 | 0.1201 | 0.4326 | 0.1661 | discussion | 0.0391 | 0.0355 | 0.0538 | 0.2963 |
| result | 0.0667 | 0.0942 | 0.2463 | 0.1288 | conclusion | 0.0219 | 0.0249 | 0.0339 | 0.2670 |
| measurement | 0.055 | 0.0688 | 0.1561 | 0.0598 | author | None | None | None | 0.1488 |
| mode | None | None | 0.1436 | None | section | 0.0552 | 0.0624 | 0.0982 | 0.1105 |
| analysis | 0.0306 | 0.0557 | 0.1195 | 0.0496 | state | None | None | None | 0.0897 |
| ratio | None | None | 0.1025 | None | point | None | None | None | 0.0618 |
| time | None | None | 0.0771 | None | mention | None | None | None | 0.0481 |
| mean | None | None | 0.0694 | None | cloud | None | None | None | 0.0379 |
| value | None | None | 0.0622 | None | findings | 0.0205 | 0.0143 | 0.0181 | 0.0373 |
| table | 0.0252 | 0.0374 | 0.0572 | 0.0363 | place | None | None | None | 0.0357 |

For answering the RQ3, we utilize four methods to analyze the correlation between the distribution of PRC in different structures and the citations. We find that the distribution of PRC in different structures of academic papers is independent of the citations. The conclusions obtained by the four methods are consistent proving that the conclusions are reliable. It can be found the distribution of PRC in different citation partitions is consistent. Materials & Methods and Results are the core parts of an academic article. It is expected that experts pay more attention to these two parts. Moreover, it could be impossible to predict the citations according to the distribution of PRC in different structures (see Figure 10).

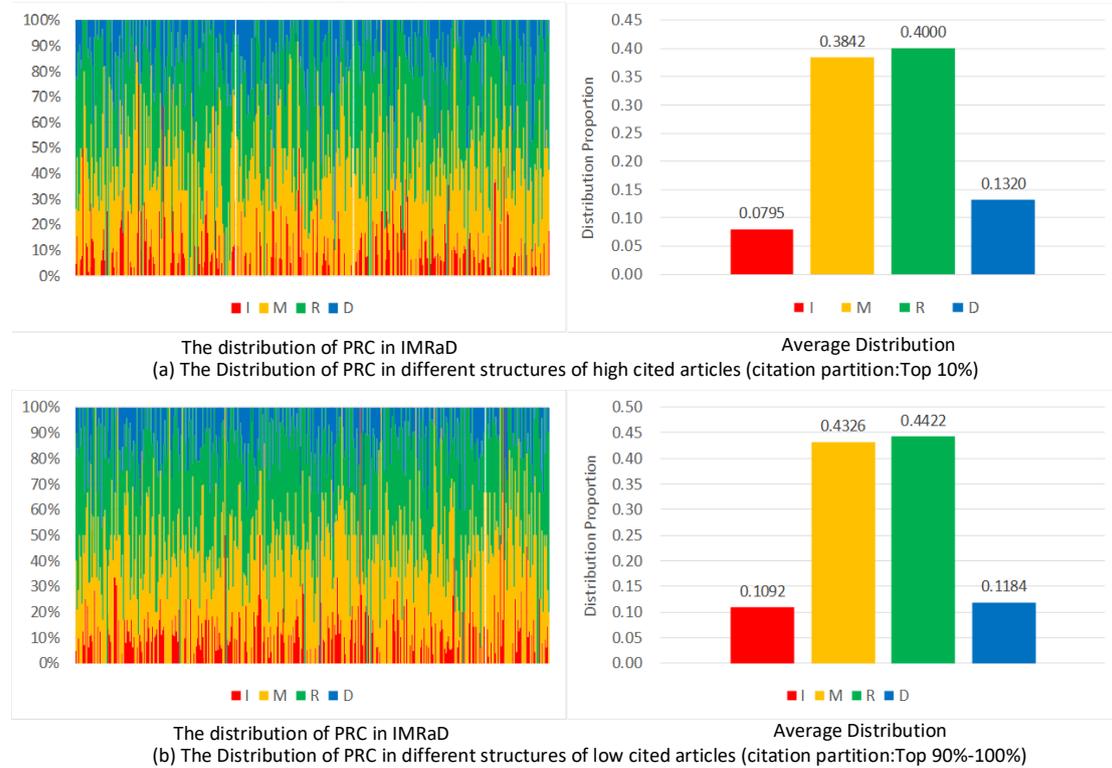

**Figure 10. The distribution of PRC in IMRaD of high cited articles and low cited articles.**

Due to the differences in the way referees write peer review reports, the rules used to extract position information cannot cover all PRC. The extraction of position information in PRC is crucial for this paper. We use the method based on rules and keyword matching to extract position information in PRC, but the coverage of location information needs further improvement, which is also a deficiency of this chapter. The main reasons are as follows: a few peer review comments do not contain position information, such as "I recommend publication" and "the author needs to proofread the grammar of the article carefully". Furthermore, there are significant differences in the writing style of different referees, and it is impossible to cover all cases. Therefore, publishing groups and journals should standardize the format of writing peer review reports to facilitate data processing better.

## Conclusions

We explore three research questions in this paper, namely, RQ1: which article structures do referees pay more attention to? RQ2: what specific content do referees focus on in different structures? RQ3: Is the distribution of PRC in different structures related to the citations? We use feature words of section title and sentence-level HAN model to identify the structure of academic papers. Then, we analyze the distribution of PRC in different structures to answer the RQ1. Secondly, we combine TF-IDF with CHI to extract the feature words of PRC distributed in different structures to answer the RQ2. Finally, we utilize four methods to analyze the correlation between the distribution of PRC in different structures and the citations to answer the RQ3. We have obtained several meaningful conclusions: the count of PRC distributed in Materials & Methods and Results is significantly more than that in Introduction and Discussion, indicating that referees pay more attention to the Material & Methods and Results; the distribution of feature words of PRC in different structures is obviously different, which can reflect the specific content of referees' concern; there is no correlation between the distribution of PRC in different structures and the citations.

This paper has other shortcomings, such as the data we used only involved one discipline, and the conclusions obtained may not be universal. However, the research methods in this paper can be generalized to other disciplines. In the future, we will focus on solving the existing deficiencies, studying the automatic identification of the types of PRC, analyzing the distribution of the types of PRC, and exploring the correlation between the types distribution of PRC in review reports and the citations. Besides, the distribution of PRC types can be combined with the structures of academic articles, and we plan to analyze the correlation between the distribution of PRC types in different structures and the citations. These explorations will help people deepen their understanding of the peer review mechanism and provide new ideas for quantifying and recognizing the contributions made by referees in the peer review process.

## Acknowledgement

This study is supported by the National Natural Science Foundation of China (Grant No. 72074113) and the opening fund of the Key Laboratory of Rich-media Knowledge Organization and Service of Digital Publishing Content (Grant No. zd2022-10/02).

## Notes

1. Peer review of nature, available at: https://www.nature.com/nature/editorial-policies/peer-review#transparent-peer-review
2. Open letter on the publication of peer review reports, available at: https://Asapbio.Org/Letter
3. FAQ: transparent peer review at nature research, available at: https://www.nature.com/documents/ nature-transparent-peer-review.pdf
4. PDF converter of Xunjie, available at: https://www.shipinzhuanhuan.cn

# Appendix A: The distribution of PRC in different structures of two dataset

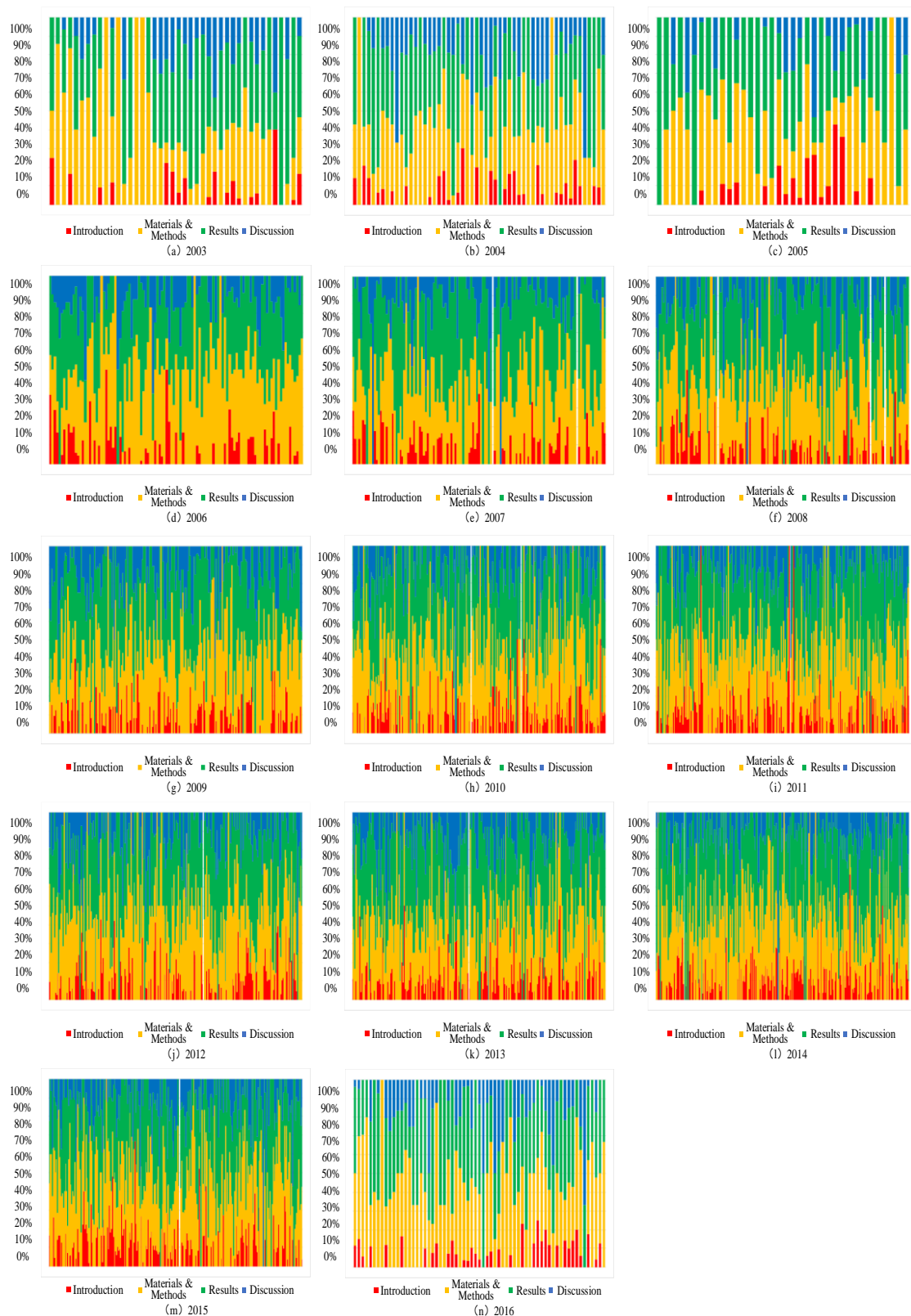

**Figure A1. The distribution of PRC in different structures in dataset-A**

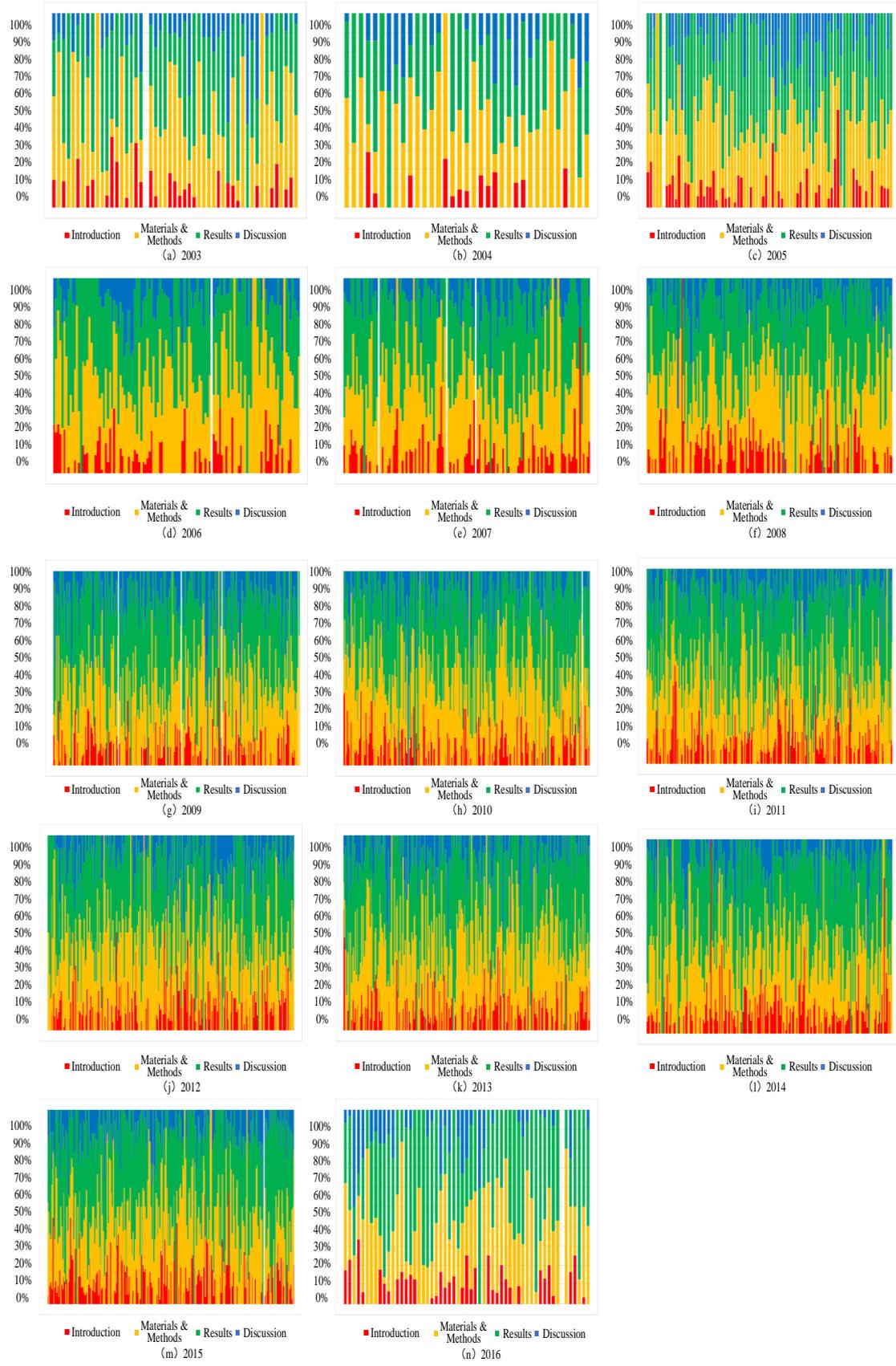

**Figure A2. The distribution of PRC in different structures in dataset-B**

# Appendix B: Normal test of the citations based on the Q-Q plot

We utilize the Q-Q plot (Quantile-Quantile plot) to test whether the citations in each year are consistent with the normal test. If the samples are normally distributed, the sample points are distributed near the standard line, approximately a straight line. The normal Q-Q plot is shown in Figure 1, which indicates that the sample points are not distributed near the standard line. Therefore, the citations do not conform to the normal distribution.

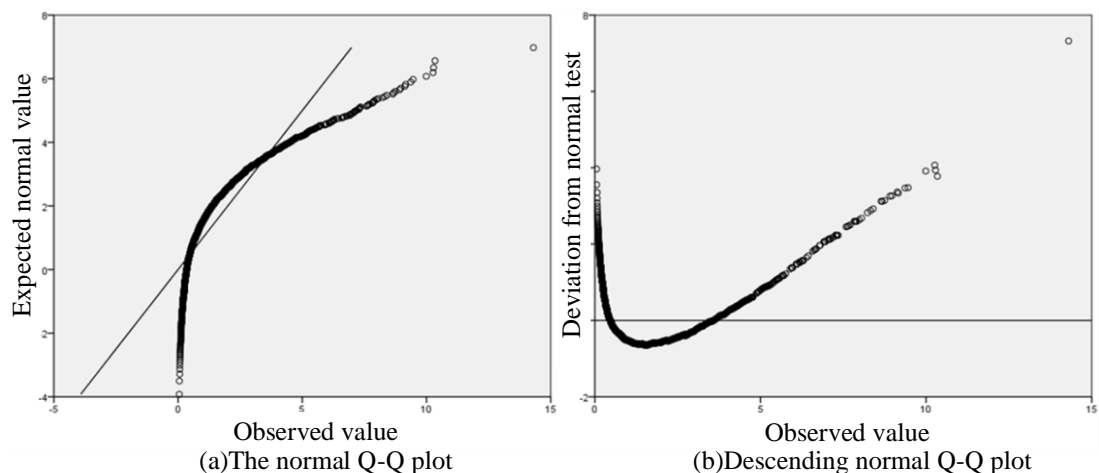

(a) The normal Q-Q plot  (b) Descending normal Q-Q plot

**Figure A3. The normal Q-Q plot of the citations of academic articles**

The K-S test is a nonparametric test method that tests whether an empirical distribution conforms to a theoretical distribution or compares whether there is a significant difference between two empirical distributions based on the cumulative distribution function. In the normal test of the citations based on the K-S test, the significance is greater than 0.05, and the null hypothesis follows a normal distribution. The result is shown in Table 1. The significance is 0.000, less than 0.05, so the distribution of the citations does not follow the normal distribution.

**Table A1. One-Sample Kolmogorov-Smirnov Test**

|  |  | Citation |
|---|---|---|
| N |  | 2,503 |
| Normal Parameters[a,b] | Mean | 1.5305 |
|  | Std. Deviation | 1.5653 |
| Most Extreme Differences | Absolute | 0.174 |
|  | Positive | 0.160 |
|  | Negative | -0.174 |
| Kolmogorov-Smirnov Z |  | 8.720 |
| Asymp. Sig. (2-tailed) |  | 0.000 |

**Note:** [*] $p<0.05$, [**] $p<0.01$, a. Test distribution is Normal, b. Calculated from data

# Appendix C: The cumulative distribution of PRC in different structures and different citation partitions

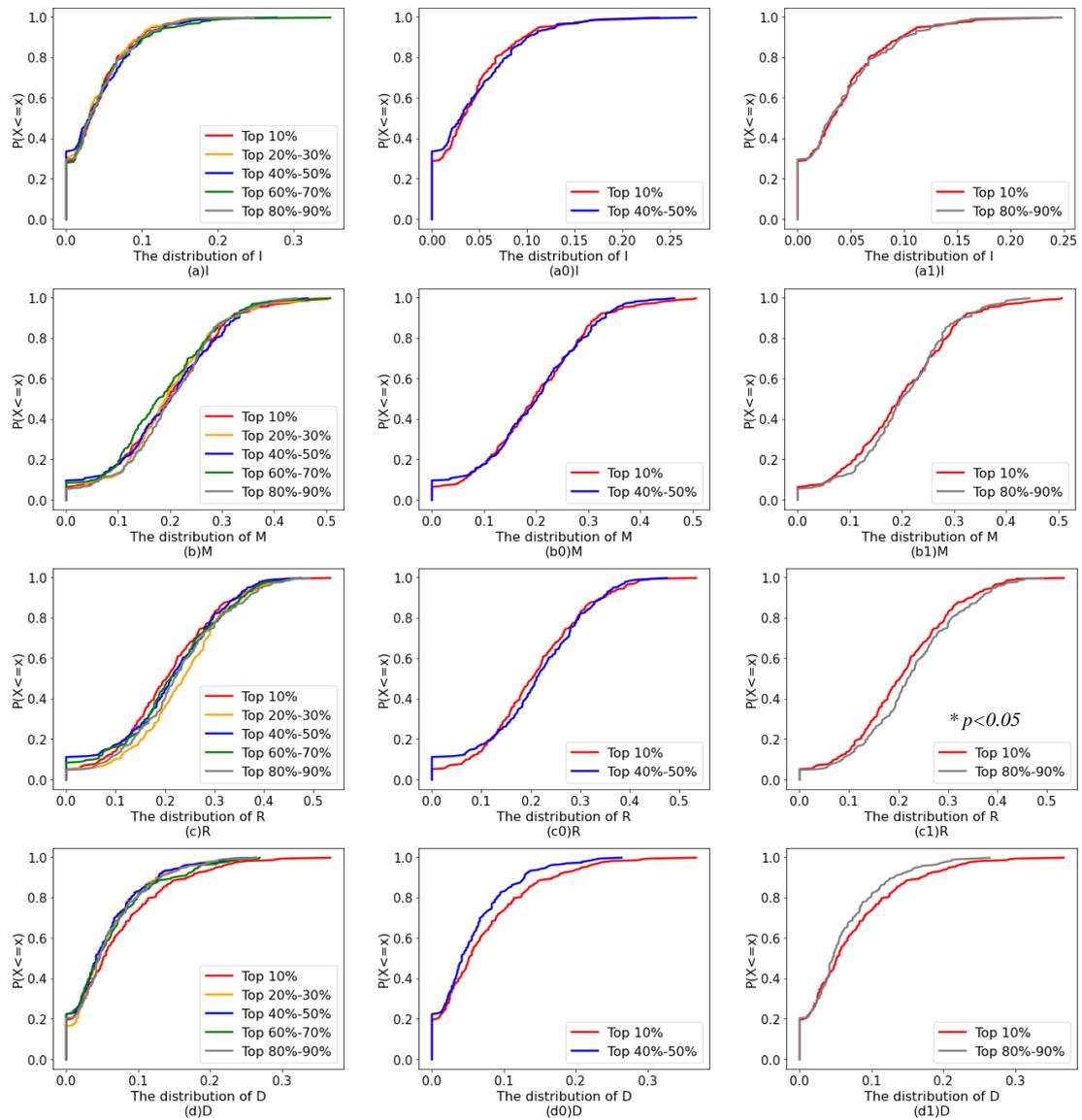

**Figure A4. The Cumulative Distribution of PRC in different structures of academic articles and different citation partitions**